\documentclass[pdftex,12pt]{article}
\pdfoutput=1  
\usepackage{arxiv}
\usepackage[utf8]{inputenc} 
\usepackage[T1]{fontenc}      
\usepackage{booktabs}       
\usepackage{microtype}      
\usepackage{graphicx}
\usepackage{bm}
\usepackage{amsmath,amsfonts,amssymb}
\usepackage{wrapfig}
\usepackage{subfigure}
\usepackage{bm}
\usepackage{natbib}
\usepackage{hyperref}
\title{Dense Object Reconstruction from RGBD Images with Embedded Deep Shape Representations}
\author{
  Lan Hu\\
  Department of Computer Science\\
  Shanghaitech  University\\
  393 Middle Huaxia Road, Pudong, Shanghai \\
  \texttt{hulan@shanghaitech.edu.cn} \\
   \And
 Yuchen Cao\\
  Department of Computer Science\\
  Shanghaitech  University\\
  393 Middle Huaxia Road, Pudong, Shanghai \\
  \texttt{caoyc@shanghaitech.edu.cn} \\
     \And
 Peng Wu\\
  Department of Computer Science\\
  Shanghaitech  University\\
  393 Middle Huaxia Road, Pudong, Shanghai \\
  \texttt{wupeng2@shanghaitech.edu.cn} \\ 
       \And
 Laurent Kneip\\
  Department of Computer Science\\
  Shanghaitech  University\\
  393 Middle Huaxia Road, Pudong, Shanghai \\
  \texttt{lkenip@shanghaitech.edu.cn} \\
}
\date{}
\begin{document}
\date{}
\maketitle
\begin{abstract}
Most problems involving simultaneous localization and mapping can nowadays be solved using one of two fundamentally different approaches. The traditional approach is given by a least-squares objective, which minimizes many local photometric or geometric residuals over explicitly parametrized structure and camera parameters. Unmodeled effects violating the lambertian surface assumption or geometric invariances of individual residuals are encountered through statistical averaging or the addition of robust kernels and smoothness terms. Aiming at more accurate measurement models and the inclusion of higher-order shape priors, the community more recently shifted its attention to deep end-to-end models for solving geometric localization and mapping problems. However, at test-time, these feed-forward models ignore the more traditional geometric or photometric consistency terms, thus leading to a low ability to recover fine details and potentially complete failure in corner case scenarios. With an application to dense object modeling from RGBD images, our work aims at taking the best of both worlds by embedding modern higher-order object shape priors into classical iterative residual minimization objectives. We demonstrate a general ability to improve mapping accuracy with respect to each modality alone, and present a successful application to real data.
\end{abstract}


\section{Introduction and review of existing work}

The ability of a machine to perceive and localize within its immediate surroundings has long been recognized as a fundamental enabler of several game-changing future technologies, including mixed or augmented reality, robotics, and intelligent vehicles. Whether we are talking about a passively moving head-mounted display in an augmented reality application or an actively navigating self-driving car, the 3D perception task remains similar:  Process the continuous input data stream from all the available sensors to solve the mutual problem of simultaneous localization and mapping (SLAM).


In an effort to include 3D geometric priors into the estimation, the community has recently also explored deep end-to-end models for the feed-forward generation of depth maps or even full object models. For example, \citep{wu2017marrnet,girdhar2016learning,yang20173d,smith2017improved} use encoder-decoder networks to output binary 3D occupancy grids from single images of an object. \citep{di2016deep,yan2016perspective,rezende2016unsupervised} furthermore train networks to perform reconstruction by utilizing multi-view images. While these models lead to surprising performance, they---at least at test time---ignore the more traditional but often valid geometric or photometric consistency constraints altogether. As a result, deep feed-forward models have difficulties to reconstruct fine details, and often fail to provide confidence measures or satisfying performance in corner-case scenarios that are insufficiently represented in the training set.

In an effort to push the performance of SLAM formulations to the next level, the community has recently investigated various strategies to combine both modalities and include prior, high-level knowledge into classical iterative residual minimization frameworks:
\begin{itemize}
\item The simplest approach of including higher-level knowledge is given by explicitly representing the common 1D or 2D geometric structures of man-made environments. Straightforward examples are given by lines and planes \citep{salas2014dense,micusik2015descriptor}. Although a well-explored idea, this technique already achieves remarkable properties that we also pursue in our work: A low-dimensional dense representation of the environment that implicitly enforces smoothness by encoding the higher-order shape of the environment.
\item Starting from these simple geometric primitives, the community has moved on to the usage of image-based semantic object detection modules. The latter typically incorporate the experience given by countless training examples into a state-of-the-art deep convolutional neural network, which is used online for retrieving a plausible CAD model to represent objects in the environment. Our work is similar in that we also leverage semantic information to change or even simplify the object representations. However, we do not rely on CAD models. While object detection modules are typically tested for generalization ability, picking a concrete model from a database for the actual SLAM optimization objective \citep{civera2011towards,salas2013slam++,gupta2015aligning,galvez2016real,mu2016slam} removes this advantage and limits accurate inference to a finite set of objects for which an exact model is known in advance.
\item More recently, the community defined the concept of \textit{semantic mapping}, which denotes the augmentation of generated 3D models by semantic information. However, methods such as \citep{koppula2011semantic,stuckler2015dense,kundu2014joint,mccormac2017semanticfusion} hardly explore the benefit of semantic knowledge in geometric inference. They merely transfer local semantic knowledge to a global representation where multiple observations from different view-points are fused\footnote{Note however that the semantic recognition part can be readily extended to use dense depth information as well.}. A notable exception is presented in \citep{haene17}, where local smoothness constraints are rendered dependent on the semantic knowledge.
\item The works closest to our approach are given by \citep{bloesch2018codeslam} and \citep{zhu2017semantic}, which utilize low-dimensional latent feature vectors to achieve low-rank representations of 2.5D depth maps and point clouds, respectively. However, \citep{bloesch2018codeslam} aims at learning a general representation for entire depth maps without employing semantic knowledge. \citep{zhu2017semantic} employs shape representations for individual objects, however only in the form of sparse point clouds for very simple shapes and without mechanisms to deal with occlusions. Furthermore, the work leaves open questions about how a shape generator mapping from a low-dimensional latent space to the full object geometry is effectively split off from the hourglass architecture taken from \citep{FanSG16}. A further notable work is given by \citep{dame13}, which however does not yet take benefit from powerful modern deep architectures.
\end{itemize}

Our contributions are as follows. We apply the idea of embedded deep shape representations to a novel RGBD incremental tracking and mapping framework for single objects. Our representation in 3D is dense and object centric, and we introduce an occlusion mask-supported strategy to actively find a trade-off between imposing priors generated by the network, and traditional residuals with respect to the measurements. We conclude with a successful application to a challenging real example, and reconstruct a dense model of a chair from real Kinect images. Section \ref{sec-concept} introduces our deep shape representation. Section \ref{sec-embedding} explains how this model is embedded into an incremental tracking and mapping framework. Section \ref{sec-experiments} finally concludes with our experimental evaluation on both artificial and real data.

\section{Object shape representation learning}
\label{sec-concept}
This section introduces our higher-level object shape representations. We start by seeing a motivation for the basic form of the representation, as well as the need for functions mapping to and from the defined shape space. We then see the exact architecture devised to realize these mappings, followed by an exposition of all details on the corresponding training procedure.

\subsection{Motivation for higher-level models}

The traditional geometric formulation of a simultaneous tracking and mapping problem starts with the definition of a measurement model. Let $\mathbf{p}_j$ be points on an object, and $\mathbf{T}_i$ the poses of viewpoints from where the object is observed. Without loss of generality, point measurements are given by
\begin{equation}
\mathbf{m}_{ij} = \pi\left(\mathbf{T}_i\mathbf{p}_j\right) + \mathbf{n}_{ij},
\end{equation}
where $\pi(\cdot)$ is a projection function that respects the intrinsic parameters of the sensor, and $\mathbf{n}_{ij}$ is an independent noise component. The residuals between the measurements and the estimated poses and points are then given by
\begin{equation}
\mathbf{r}_{ij} = \mathbf{m}_{ij} - \pi\left(\mathbf{T}_i\mathbf{p}_j\right).
\end{equation}
A popular alternative to solve the tracking and mapping problem is to search for poses and points that minimize the sum of least squares residuals. However, in order to improve the conditioning of the problem, the energy is often complemented by a dual smoothness term that enforces neighbouring points to remain close to each other. The final objective may read
\begin{equation}
\{\mathbf{T}_{i,opt},\mathbf{p}_{j,opt}\}=\underset{\mathbf{T}_i,\mathbf{p}_j}{\operatorname{argmin}}\left[\sum_{i}\sum_{j}\mathbf{1}_{ij}\eta(\mathbf{r}_{ij}) + \sum_{(i,i')\in\mathcal{N}_{\mathbf{p}}}\eta'(\mathbf{p}_i - \mathbf{p}_{i'})\right],
\end{equation}
where $\mathbf{1}_{ij}$ is an indicator function that indicates whether point $j$ is visible in view $i$, $\mathcal{N}_{\mathbf{p}}$ is the set of index-pairs for points that are defined to be neighbors, and $\eta(\cdot)$ and $\eta'(\cdot)$ are robust cost functions. There are several problems with this objective. The first one is that the dimensionality of the problem is potentially very high, especially if we are considering the dense scenario. Then, since not all residuals are sensitive w.r.t. the camera pose, the correct solution depends on the additional smoothness term. Even though it can be solved efficiently via a primal-dual method, it does raise the complexity of the optimization problem. The final problem of the formulation is that the measurement model is very simple, and does not take gross errors caused by reflections or complicated illumination conditions into account. The robust kernels and the smoothness term have only limited ability to deal with such effects.

We now assume that we have a compact way to describe the object's shape given by an $n$-dimensional shape descriptor $\bm{\lambda}$ and a function to map from the latent shape space to the full 3D geometry. Let $\mathcal{P}(\bm{\lambda})$ be the set of points describing the object's shape and generated by $\bm{\lambda}$. The objective of tracking and mapping may now be reformulated as
\begin{equation}
\{\mathbf{T}_{i,opt},\bm{\lambda}_{opt}\}=\underset{\mathbf{T}_i,\bm{\lambda}}{\operatorname{argmin}}\sum_{i}\sum_{j}\mathbf{1}_{ij}\eta( \mathbf{m}_{ij} - \pi\left(\mathbf{T}_i\mathcal{P}(\bm{\lambda})[j]\right) ),
\label{eq-embedded_shape_model}
\end{equation}
a much simpler formulation that has a few encouraging properties such as low-dimensionality, implicit smoothness in the generated point cloud, and---assuming that the shape representation is strong enough to only generate valid shapes---a high resilience against effects that are not modeled by simple measurement functions.

\subsection{Architecture}

\begin{figure}[b!]
  \centering
  \includegraphics[width=0.8\textwidth]{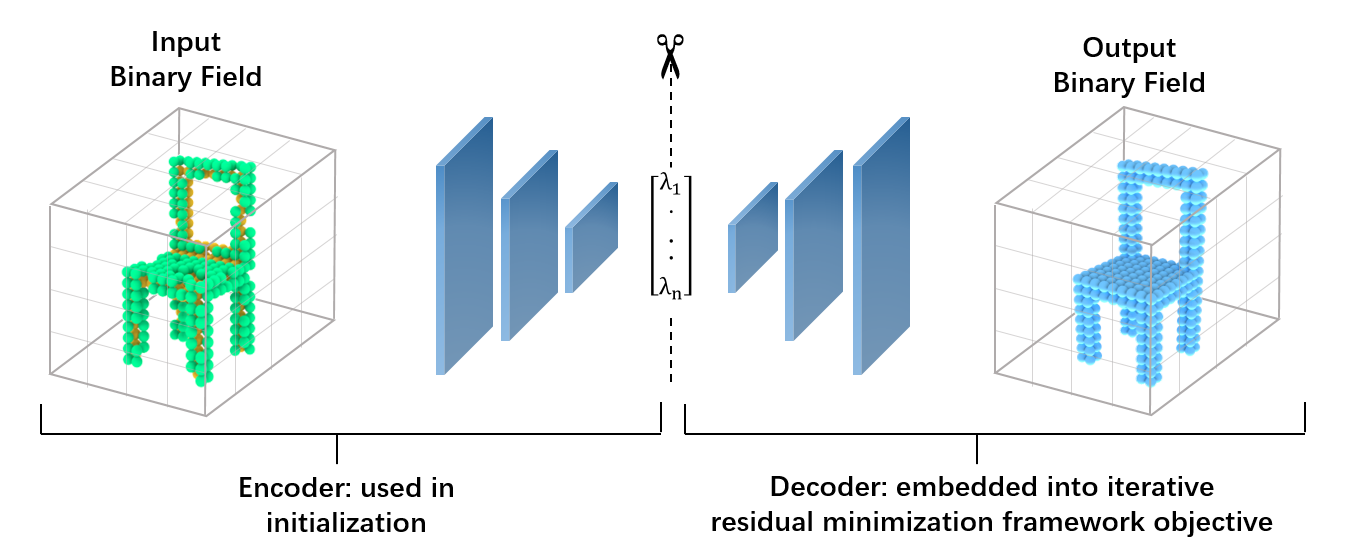}
  \caption{Inspired by \citep{girdhar2016learning}, we use a deep auto-encoder to train a generic low-dimensional shape representation for a certain class of objects. After training is completed, encoder and decoder are separated and used for initialization and iterative residual minimization, respectively.}
  \label{fig-autoencoder}
\end{figure}

There are various ways to formulate the low-dimensional shape representation such as PCA or LLE, but we choose here auto-encoders which have proven to be potentially very good at this task~\citep{girdhar2016learning}. As indicated in Figure \ref{fig-autoencoder}, we structure the object shape in the original space with a binary voxel grid, and train an auto-encoder to reproduce a full model of the object. Unlike \citep{wu2017marrnet,smith2017improved,yan2016perspective,rezende2016unsupervised}, which generate the full $3D$ shape from the RGBD image directly, the voxel grid fuses the information from different views more efficiently, and also does not limit the number of views. We only use simple occupancy information for map encoding, where 1 represents an occupied cell, and 0 an empty one. All our grids (i.e. the input which is the measurements, denoted $\mathbf{F}$, the output, denoted $\mathbf{G}$, and the ground truth 3D shape, denoted $\mathbf{Y}$) are $32\times 32\times 32$ occupancy grids. 

In this work, we focus on the example object class of chairs, which we believe to be interesting since sharing commonalities while at the same time having relatively complex shape and intra-class shape variations. Note that---after training is finished---we may separate the encoder from the decoder, and thus obtain our mapping function $\mathcal{P}(\bm{\lambda})$ from the latent shape space to the full geometry that we would like to embed into a residual minimization framework. This is particularly supported by the fact that we do not employ any skip connections in the network.

Besides obtaining a low-dimensional shape representation, we also want our network to learn how to predict the full $3D$ shape from only partial input binary fields. By doing so, we enable our network and in particular the encoder part to initialize the latent shape descriptor directly from our fused measurements $\mathbf{F}$ (i.e. one or several fused $2.5D$ depth images\footnote{Note that, once camera poses are known, a partial 3D binary field is readily obtained from 2.5D depth images by recalculating 3D points and finding intersecting voxels.}). However, note that we still limit ourselves to the intrinsic object shape, and therefore only use similarly oriented chairs. The geometry of Euclidean poses is well understood, and explicitly parametrized in our formulation.

\begin{figure}[t]
  \centering
  \includegraphics[width=\textwidth]{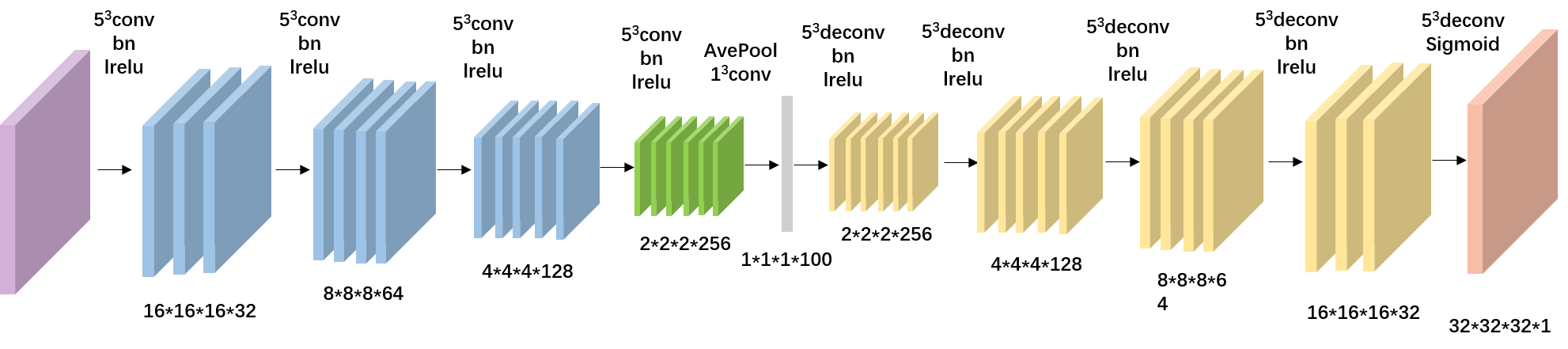}
  \caption{Detailed architecture of our encoder-decoder network.}
  \label{fig-autoencoder_structure}
\end{figure}

The dimensionality of the shape representation is the most important design factor. Choosing a high dimensionality may generate many saddle points in the shape space \citep{dauphin2014identifying}, which makes it difficult for our subsequent iterative residual minimization scheme to converge to the optimal solution. While \citep{yang20173d}, \citep{dai2017shape}, and \citep{wu2016learning} use more than $200$ dimensions, we managed to reduce the number to $100$, and thus smooth out the topology of our optimization space.

The detailed structure of our auto-encoder neural network is given by the encoder and decoder illustrated in Figure \ref{fig-autoencoder_structure}. Our encoder is a down-sampling network, encoding a $32\times 32\times 32$ binary grid into the latent shape representation with a dimensionality of $100$. There are five convolution layers. The first four layers have a similar structure, each of them applying a bank of $5\times 5\times 5$ convolution filters with stride $2$, followed by a $3D$ batch normalization and a leaky relu activation function. The fifth layer is made up by 3D average pooling and a 3D convolution with $1\times 1\times 1$ filters and stride $1$, which are used to substitute the full linear transformation to avoid over-fitting and sustain spatial information \citep{lin2013network}. The function of the decoder is to generate the full $3D$ shape from the latent shape representation. Our decoder has four  $5\times 5\times 5$ deconvolution layers with stride $2$, followed by a $3D$ batch normalization and a leaky relu activation function except for the last layer, which has a tanh activation function. Due to the range of the output, we add a linear transformation mapping from $(-1,1)$ to $(0,1)$ after the tanh function.


\subsection{Training}
\label{sec-training}

We use $V_{ijk}$ to represent the value at position $(i,j,k)$ in a 3D voxel grid $\mathbf{V}$. The loss function of the network is given by the binary cross entropy
 \begin{equation}
 L_{\text{ae}}= \frac{1}{N^3}\sum_{i}^{N}\sum_{j}^{N}\sum_{k}^{N} -Y_{ijk} \text{log}(G_{ijk})-(1-Y_{ijk})\text{log}(1-G_{ijk}),
 \end{equation}
where $N$ is the resolution $32$, $Y_{ijk}$ the target value in $\{0,1\}$, and $G_{ijk}$ the estimated value in $(0,1)$. We derive our training data from CAD chair rendering models \citep{chang2015shapenet}. The input shapes are given by partial voxel grids obtained from depth images. We therefore start by hypothesizing a virtual depth camera scanning frames from $18$ different view-points for each CAD model. The altitude and azimuth of the views randomly ranges from $(0, 360)$ and $(30, 60)$, respectively. The principal axis of the views are intersecting with the object center, and the $x$-axis of each view remains horizontal. Virtual depth images are generated by projecting each triangle into the image, finding the intersecting pixels, and intersecting the corresponding rays with the triangle in 3D to recover the depth. A depth check is added to handle occlusions. 
\begin{figure}[t!]
  \centering
  \includegraphics[width=1\textwidth]{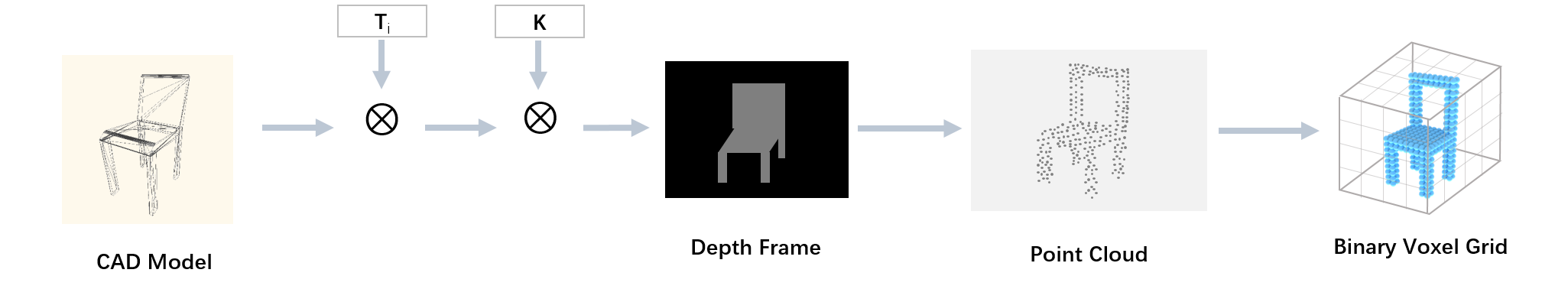}
  \caption{Processing pipeline for generating artificial partial binary grid scans.}
  \label{fig-render}
\end{figure}
Once a depth image is generated, it is transformed back into a 3D point cloud using the virtual camera parameters.
The last step then consists of transforming all point clouds into binary voxel fields, which is easily achieved by setting all the voxels that contain a 3D point to $1$, and the rest to $0$. Note that---in order to make sure the ground truth example is really complete---the point cloud used to generate the ground truth binary field is derived straight from the CAD model without bypassing via the synthesized depth frames. Also note that the transformation from the point clouds to the binary fields first employs anisotropic scaling and a translational shift to make sure the grid resolution is ideally exploited. For each training example, we generate these transformation parameters only once straight from the CAD model, and then apply them without change to the partial views as well. We obtain $18$ pairs of partial and complete voxel grids for each chair and use $3858$ different chairs for training our network, which means $69444$ training pairs in total.

We initialize all the convolution and normalization parameters of the network by sampling from the gaussian distributions $(0, 0.02)$ and $(1, 0.02)$.  The parameters of the Adam solver are $\beta_1= 0.9$ and $\beta_2 = 0.999$, and the learning rate is set to $0.0001$ for $200$ batches. The entire network is trained from scratch with two 1080Ti GPU using Pytorch.

\section{Embedding into Iterative Residual Minimization}
\label{sec-embedding}

\begin{figure}[t]
 \centering
 \includegraphics[width=\textwidth]{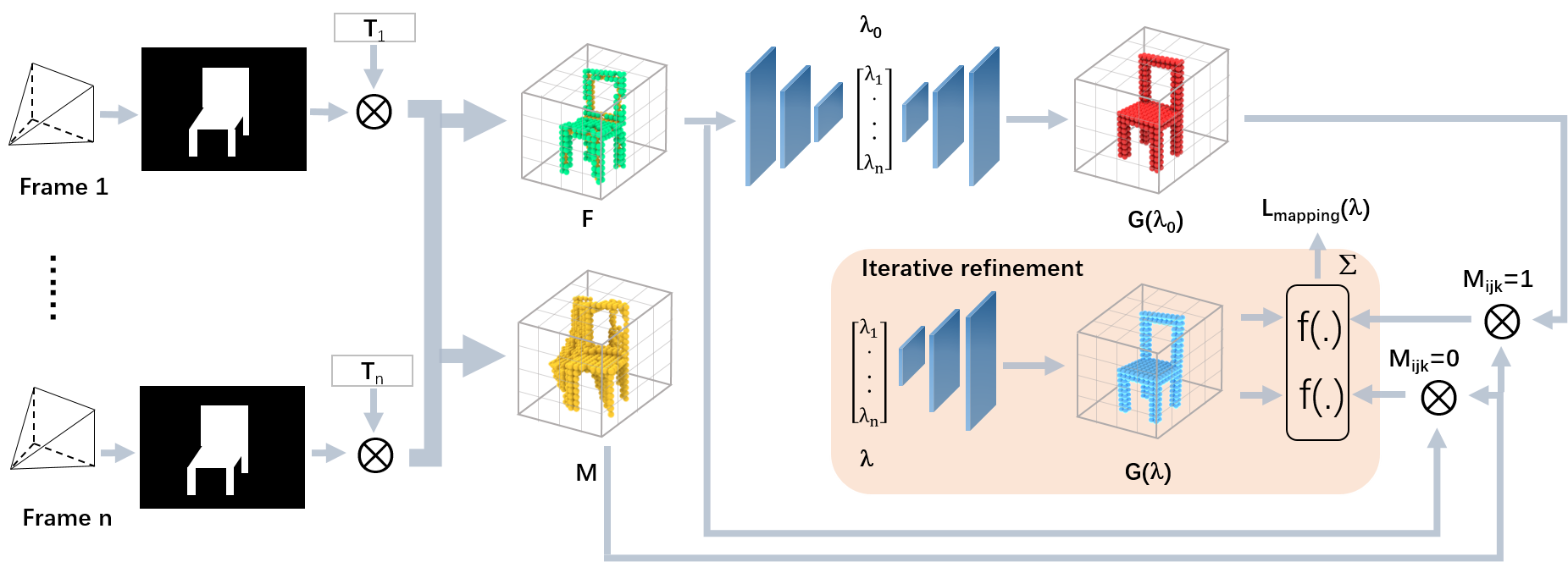}
 \caption{Overview of our shape prior based reconstruction pipeline. After segmentation, the frames are first compensated for their pose and fused to generate an occupancy measurement $\mathbf{F}$ as well as an occlusion mask $\mathbf{M}$. $\mathbf{F}$ is sent into the auto-encoder network to generate the shape prior $\mathbf{G}(\bm{\lambda}_0)$. Then, $\mathbf{F}$, $\mathbf{G}(\bm{\lambda}_0)$, and $\mathbf{M}$ are  used to construct cross-entropy residuals with respect to the output of the network. Finally, these residuals are iteratively minimized over the latent shape representation $\bm{\lambda}$.}
 \label{fig-pipeline}
 \end{figure}

This section illustrates how we iteratively refine the generated $3D$ shape by embedding the model into a traditional residual minimization framework. The computation is divided into two stages. The first one involves the generation of a 3D shape prior by a standard feed-forward execution of the network. The second part then consists of the iterative refinement during which the consistency with the original measurements is taken into account. The section concludes by a brief exposition of our incremental, alternating tracking and mapping scheme. Figure \ref{fig-pipeline} illustrates a break-down of our complete pipeline.

\subsection{Shape prior generation}

We start by a plain feed-forward application of the entire auto-encoder network to initialize our latent object shape representation and obtain a prior on the full 3D geometry. With respect to Figure \ref{fig-pipeline}, this is the top part of the flow-chart. The procedure works as follows. We first start by applying an object segmentation in each RGBD frame (on synthetic datasets, this step is not necessary, and on real datasets, we simply perform a RGB-image based foreground-background segmentation). Assuming that the pose of each RGBD frame has already been identified, we then take the 3D points measured on the object's surface and transform them into the world frame (which coincides with the object frame in our work). How to estimate and gradually update the camera pose of each frame will be discussed in Section \ref{sec-poserecov}. Similarly to the training dataset generation, we complete the input generation by estimating isotroping scale factors from the 3D points, and then transform each partial point cloud into a binary occupancy grid using the discretization function $D(.)$. If $\mathbf{P}_{i}$ denotes the $3\times x$ matrix of 3D points observed in frame $i$, the fused measurement $\mathbf{F}$ is finally given as
\begin{equation}
\mathbf{F} =  D(\mathbf{T}_1\mathbf{P}_1) \lor D(\mathbf{T}_2\mathbf{P}_2)  \lor \cdots \lor D(\mathbf{T}_n\mathbf{P}_n),
\label{eq-fusion}
\end{equation}
where $\lor$ is the element-wise OR operation between two voxel grids. $\mathbf{F}$ is the input for our auto-encoder predicting the full $3D$ shape $\mathbf{G}(\bm{\lambda}_0)$, where $\bm{\lambda}_0$ is our initial low-rank shape representation.

\subsection{Iterative shape refinement}
We now proceed to the core of our contribution. Most existing approaches employing shape priors do not subject the network's prediction to any further post-processing. The prediction of the network however tends to be blurry and often misses out on fine structure details. It furthermore comes without any guarantees and---as shown in numerous works---networks can indeed be ``fooled'' by inputs that otherwise would seem to be relatively normal examples. We therefore construct residuals of the network output with respect to the original measurements, and iteratively minimize those residuals as a function of our latent shape representation $\bm{\lambda}$. Note that, although the introduction of binary occupancy grids puts this into a different form, the basic idea of this approach is similar to the one presented in (\ref{eq-embedded_shape_model}), namely a shape generator embedded into traditional residuals.

\begin{wrapfigure}{r}{0.4\textwidth}
\vspace{-1cm}
\includegraphics[width=0.4\textwidth]{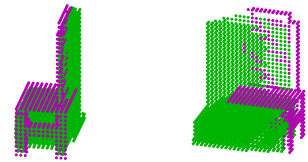}
\caption{Visualization of the mask $\mathbf{M}$. Purple points are on the surface. Checking the distance of each voxel from the camera center and comparing against the measurements then defines free space (invisible) and occluded points (in green).}
\vspace{-0.5cm}
\label{fig-show_mask}
\end{wrapfigure}
Using only the measurements $\mathbf{F}$ to construct these residuals however bares a trap. Not every point in space may have been visible, and parameters defining the shape in unobserved regions would become fully unconstrained (recall that the encoder is no longer employed during residual minimization). Our optimization needs to be further constrained on the prior $\mathbf{G}(\bm{\lambda}_0)$, especially in the unobserved regions. Finding a good balance is however difficult, as we still want to remain able to exploit the details that are potentially captured by $\mathbf{F}$ in the observed parts of space.

To solve this problem, we introduce the occlusion mask $\mathbf{M}$, which indicates which points in space have not been observed by any previous measurements. The mask also has a voxel grid format of similar resolution as $\mathbf{F}$ and $\mathbf{G}$. As explained in Figure \ref{fig-show_mask}, $\mathbf{M}$ can easily be constructed by a simple heuristic that sets voxels on the surface as well as all voxels with smaller depth from the camera center than the one measured at the reprojection location to zero, and leaves all other points equal to one (i.e. the occluded parts). Note that masks from different views can be combined  by element-wise AND operation. 

The mask allows us to balance between the measurements $\mathbf{F}$ and the prior information $\mathbf{G}(\bm{\lambda}_{0})$ to update our shape representation, especially in the occluded regions. Our objective for optimizing the structure is finally given by
\begin{equation}
\begin{aligned}
L_{\text{mapping}}(\bm{\lambda})
= & \small{\sum_{i}} \small{\sum_{j}}  \small{\sum_{k}}  ~\Big \{ (1-M_{ijk}) f\big(G_{ijk}(\bm{\lambda}),F_{ijk}\big) \\
&+ \alpha M_{ijk} f\big(G_{ijk}(\bm{\lambda}),G_{ijk}(\bm{\lambda}_0)\big) \Big\}
\label{eq-mapping_loss}
\end{aligned}
\end{equation}
where $f(\cdot)$ is the binary cross entropy function already introduced in Section \ref{sec-training}, and $\alpha$ is an overall trade-off factor defined as $\alpha = \frac{ \sum_{i}\sum_{j}\sum_{k} M_{ijk} }{ \sum_{i}\sum_{j}\sum_{k} (1-M_{ijk})}$. $\alpha$ governs the overall amount at which we enforce the prior in occluded regions. Especially for the first few frames, the measurement coverage may not be very high, thus making the regularization on the prior more important. Once more frames have been accumulated, the measurements themselves are typically strong enough to regularize the latent shape representation $\bm{\lambda}$.
 
\subsection{Optimization}

We explored two strategies for refining the latent shape representation based on minimizing (\ref{eq-mapping_loss})-
\begin{itemize}
\item Gradient descent- The partial derivatives of (\ref{eq-mapping_loss}) w.r.t. $\bm{\lambda}$ are readily computed by applying gradient back-propagation in Pytorch. It is however difficult to optimize the shape as the decoder represents a highly-nonlinear function, and the optimization problem thus turns non-convex. Local optimization methods are thus prone to get trapped in local minima and saddle points.
\item Covariance Matrix Adaptation Evolution Strategy (CMA-ES)- CMA-ES is a search method known for its state-of-the-art performance in derivative-free optimization of non-linear or non-convex optimization problems \citep{hansen2016cma}. In contrast to most classical methods, fewer assumptions are made on the underlying objective function, and it can easily be applied to a black-box decoder network for which derivatives are very hard to compute. As shown in Figure \ref{fig-gm}, gradient descent converges smoothly, which means it is easy to get trapped in the nearest local extremum or saddle point.  CMA-ES in contrast declines in a fluctuating fashion, indicating that it is much better at overcoming local minima. Further comparisons between gradient descent and CMA-ES are given in Section \ref{sec-experiments}.
\end{itemize}

\subsection{Incremental tracking and mapping} 
\label{sec-poserecov}

So far we have assumed that the poses are simply given, which is unrealistic. In practice, the pose of a newly arriving frame is simply initialized by running the ICP algorithm w.r.t. to previous frames \citep{besl1992method}. The frame is added to our set of \textit{keyframes} if sufficient displacement has been detected. Each time the set of \textit{keyframes} is incremented, we fuse new grids for the measurements and the occlusion mask, and rerun our mapping paradigm. Keyframes can furthermore be realigned with the mapping result in an alternating fashion, which removes the global drift.

 
\section{Experiments}
\label{sec-experiments}

\begin{figure}
\begin{minipage}[t]{0.5\linewidth}
\centering
\includegraphics[width=0.6\textwidth]{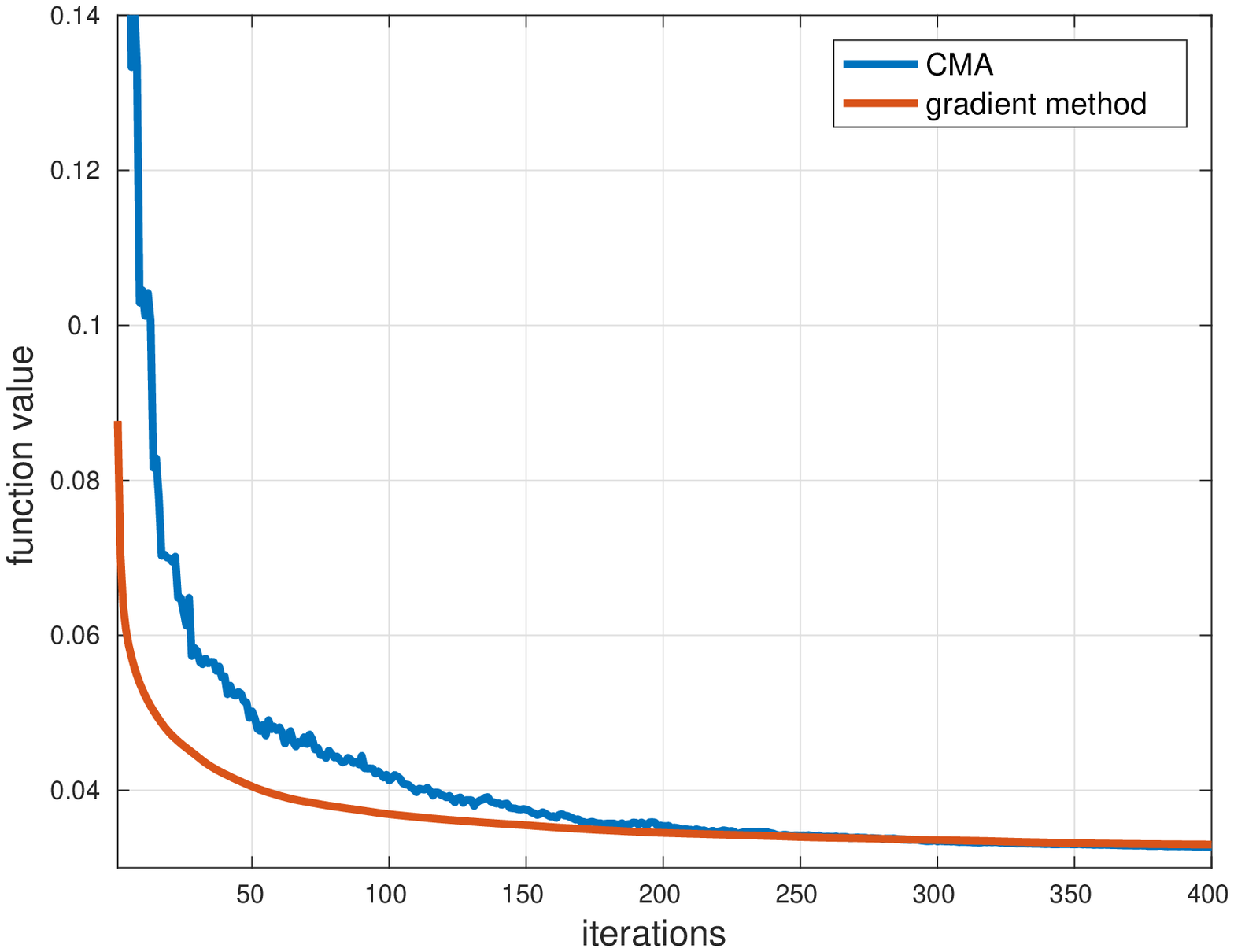}

\end{minipage}%
\begin{minipage}[t]{0.5\linewidth}
\centering
\includegraphics[width=0.6\textwidth]{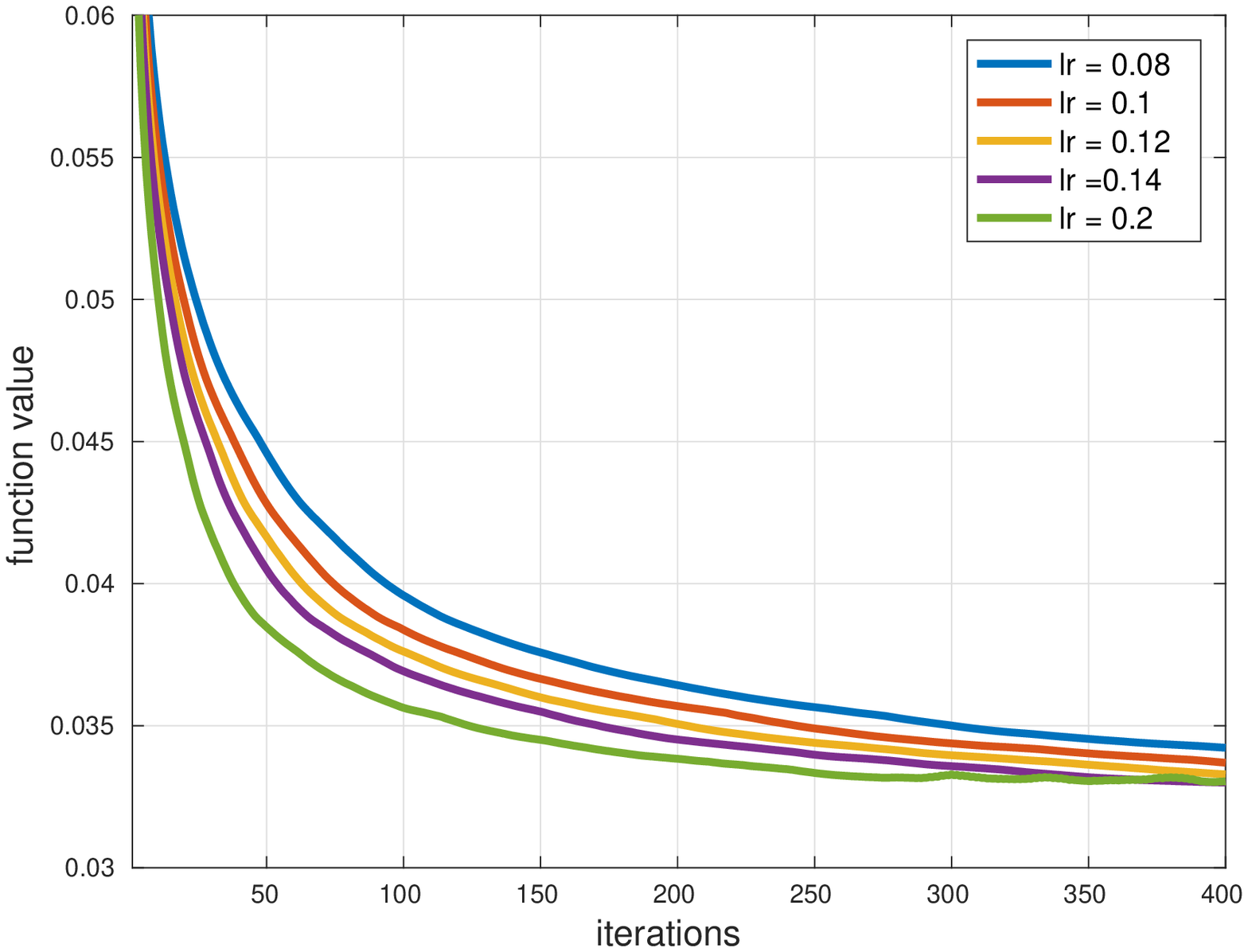}
\end{minipage}
  \caption{The left figure shows the convergence behavior of CMA-ES and gradient descent. The right one shows the convergence of gradient descent for different step sizes.}
  \label{fig-gm}
\end{figure}

We evaluate our method qualitatively and quantitatively on both synthetic data and real data. We start by tuning the step size of gradient descent in a dedicated offline experiment. As illustrated in Figure \ref{fig-gm}, averaging the residual behavior over many different iterative minimization procedures and for different step sizes reveals that a value of $0.14$ leads to fast convergence without overshooting. The starting point of gradient descent is always set to $\bm{\lambda}_{0}$. CMA-ES is implemented by the python package \emph{Pycma}. The initial mean vector is again set to $\bm{\lambda}_{0}$, and the initial covariance is set to $2$ (the value has again been identified by a dedicated experiment). The maximum number of iterations is set to 350 for both methods. To make quantitative statements about the quality of the reconstruction, we evaluate the Intersection-over-Union(IoU) between the predicted and the ground truth 3D binary occupancy grid, which is defined as follows-
\begin{equation}
\text{IoU} = \frac{\sum_{i}\sum_{j}\sum_{k} \big[I(G_{ijk}(\bm{\lambda})>p) \cdot I(Y_{ijk})\big]}{\sum_{i}\sum_{j}\sum_{k}\big[I(G_{ijk}(\bm{\lambda})>p)+I(Y_{ijk})\big]}
\end{equation}
where $I(\cdot)$ is  an indicator function, and $p$ is set to $0.5$.

\subsection{Iterative refinement on synthetic data}

\begin{figure}[h]
\begin{minipage}[t]{0.5\linewidth}
\centering
\includegraphics[height=1.3in]{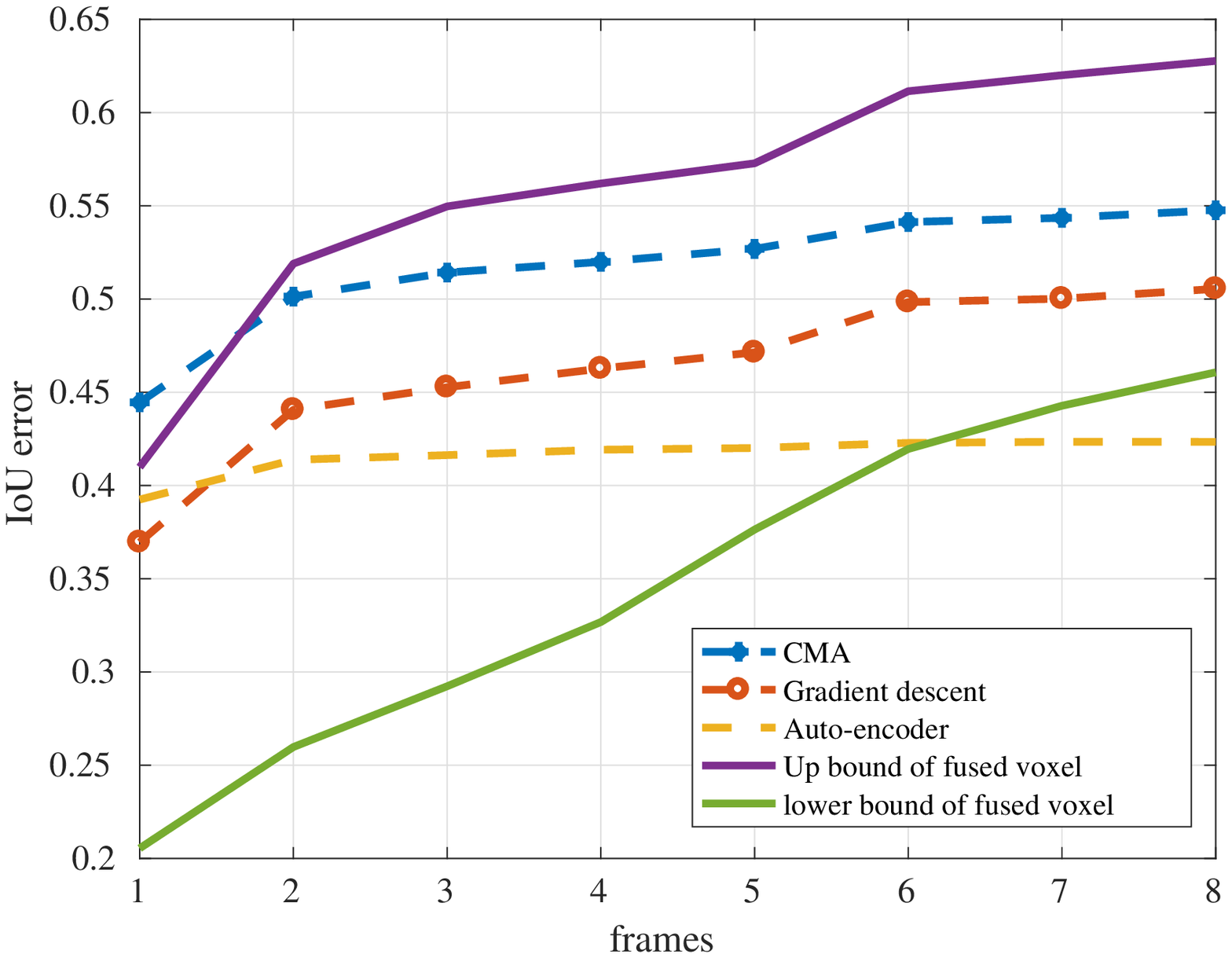}
 \caption{IoU of pure auto-encoder prediction and CMA-ES and gradient descent based refinements.}
  \label{fig-IoU-error}
\end{minipage}%
\begin{minipage}[t]{0.55\linewidth}
\centering
\includegraphics[height =1.1in]{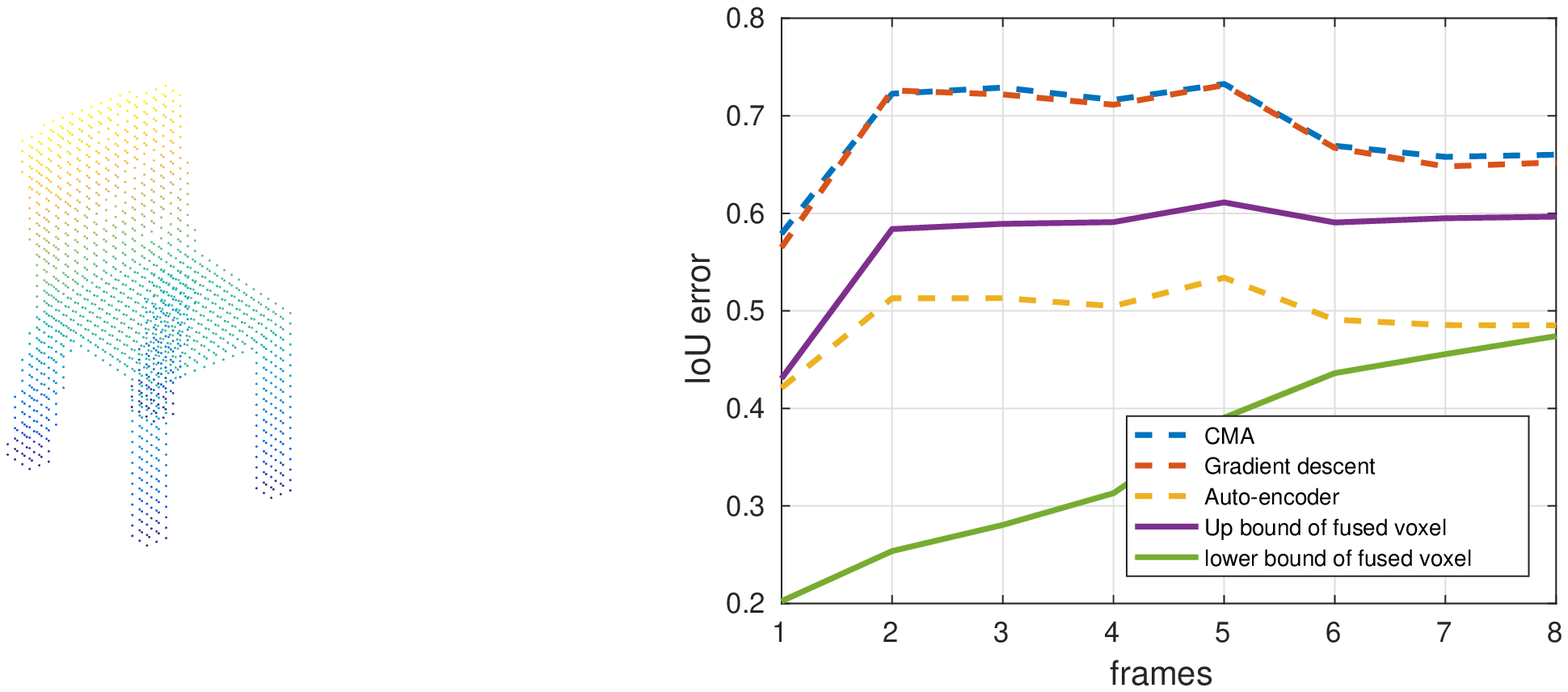}
\caption{IoU error for voluminous chairs}
\label{fig-true_false}
\end{minipage}
\end{figure}

We render synthetic test cases again by using CAD models of chairs from ShapeNet. However, unlike the test cases generated for training our network, here we always generate a continuous stream of images captured along a virtual orbit around the chair. This allows us to test the incremental mapping performance of our paradigm. Our results on synthetic data focus on the mapping part, hence the positions and orientations of each frame are fixed to their original values. We evaluate the plain auto-encoder prediction and the refined results after gradient descent and CMA-ES. Figure \ref{fig-IoU-error} shows the IoU results obtained for the different approaches and notably averaged over $288$ different random chairs.

Our experiments show that CMA-ES based iterative refinement performs better than the auto-encoder alone, especially as the number of frames is increasing. The performance of the auto-encoder is simply limited to what has been covered by the training set. The two additional solid lines in Figure \ref{fig-IoU-error} show an upper and a lower bound of the IoU error obtained by additionally setting all occluded voxels in the measurement grid to either 0 or 1, respectively. Owing to the sparse structure of the chair, most of the unobserved parts are indeed empty, so the upper bound turns out to be very accurate.(Note- if repeating the same experiment with more voluminous chairs, the upper-bound quickly becomes worse than the CMA output, the result show in Figure \ref{fig-true_false}.)

\begin{figure}[b!]
 \centering
\includegraphics[width=\textwidth]{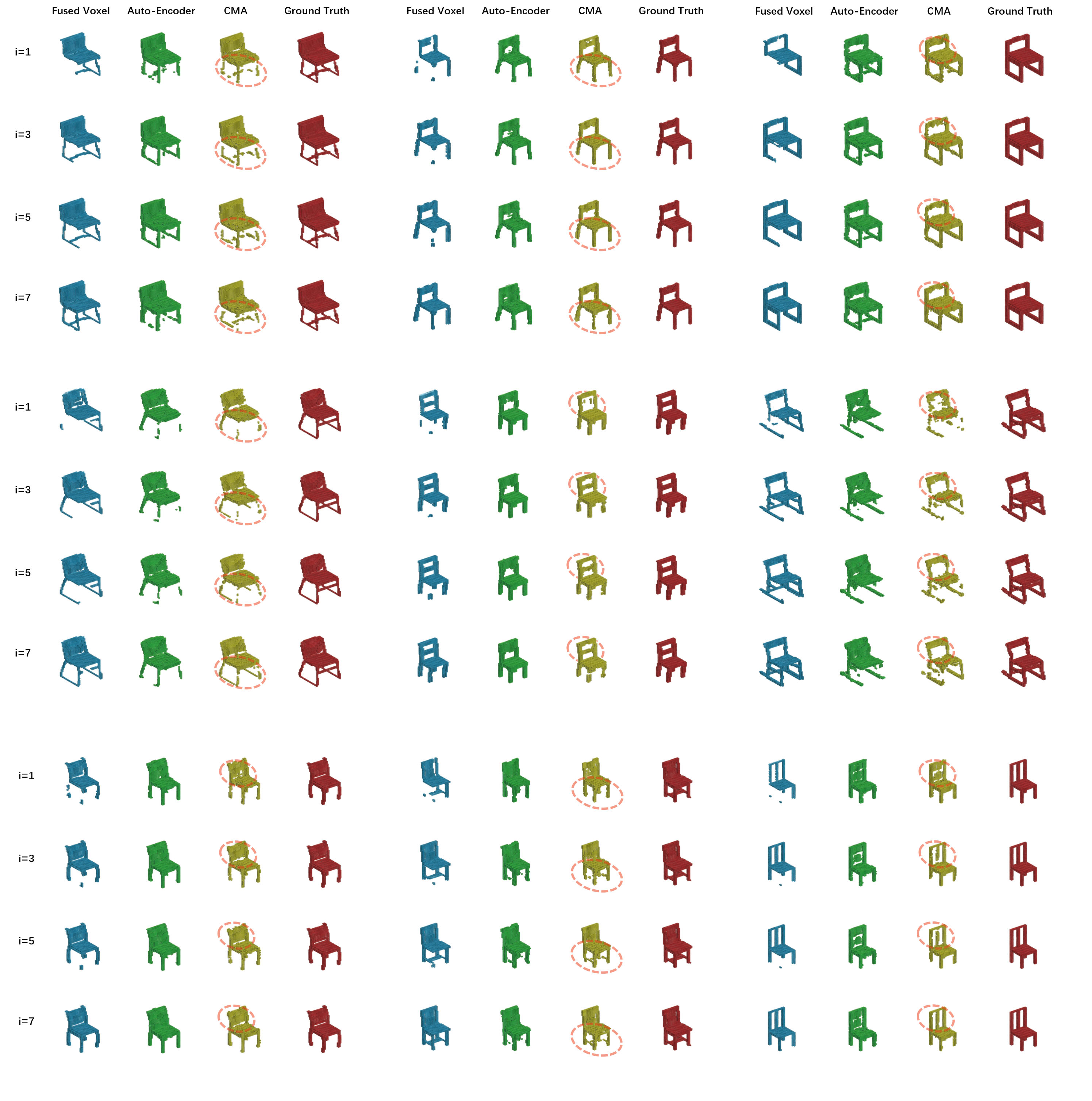}
 \caption{Visualization of CMA-ES based refinement result (in yellow) . Most of the reconstructed results exceed the results of the auto-encoder (in green) and are able to gradually approach ground truth (in red). $i$ denotes the number of frames taken into account. The blue column denotes the fused measurement grid $\mathbf{F}$ suffering from occlusions. Fine details recovered through embedding into iterative residual minimization are highlighted.}
 \label{fig-mapping}
\end{figure}

\begin{figure}[t!]
 \centering
  \includegraphics[width=0.8\textwidth]{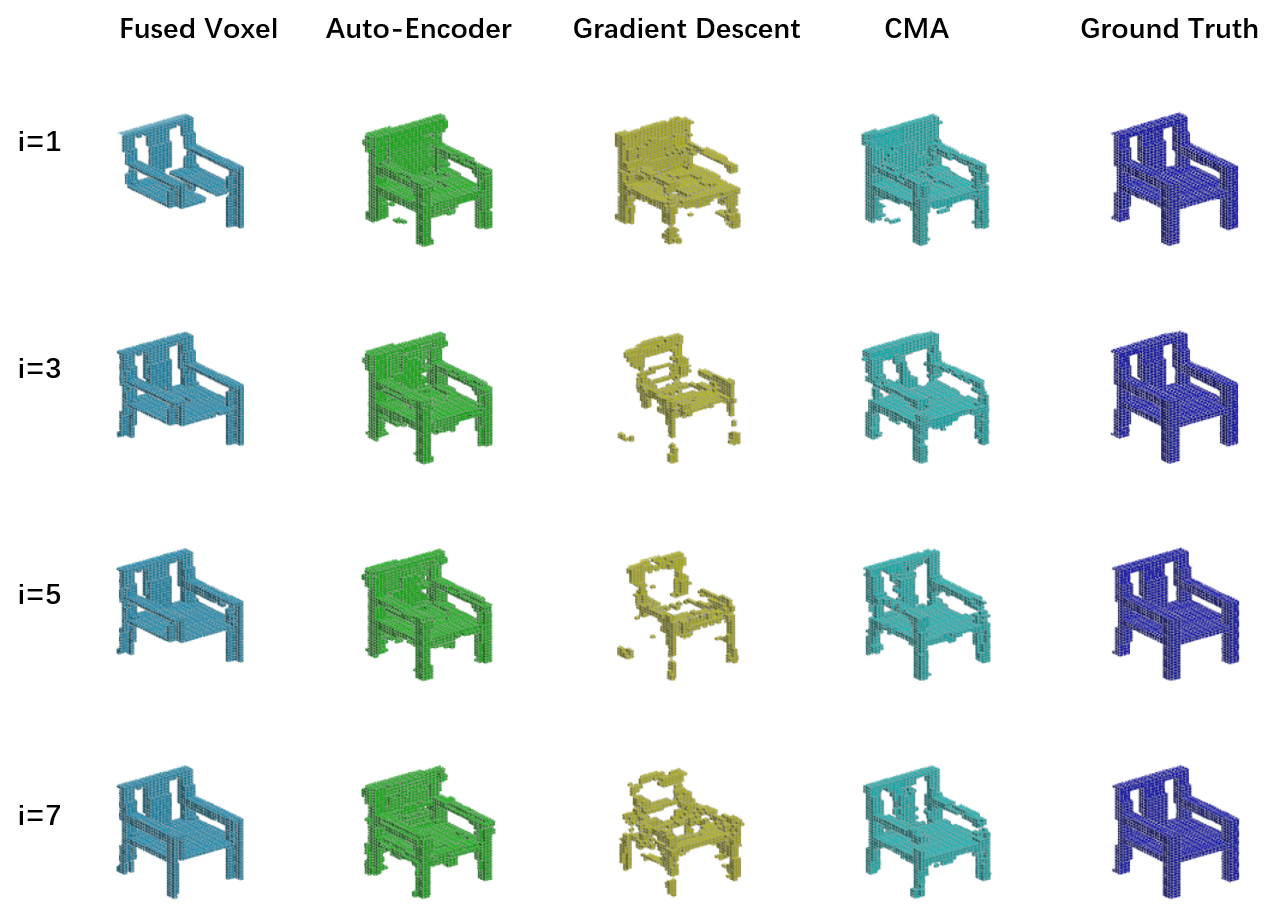}
  \caption{Illustration of failure cases of gradient descent. While CMA-ES (fourth column) is able to use additional frames to eventually approach groundtruth (and outperform the auto-encoder), gradient descent (third column) gets trapped in a local minimum early on.}
  \label{fig-gm_fail}
\end{figure}

Note however that in the general case there is no simple rule to define the value of the unobserved voxels, which is why the result from CMA-ES has to be interpreted as the  best result. Figure \ref{fig-mapping} shows some visual results comparing the output of CMA-ES against the prior from the auto-encoder and ground truth. It can be concluded that the refined model provides more detail, especially for fine structures on the legs and the back of the chair. Figure \ref{fig-gm_fail} furthermore explains the reason why---in average---gradient descent cannot outperform CMA-ES. Gradient descent simply converges to the nearest local minimum or saddle point, a potentially wrong solution. Compared to CMA-ES, it consequently has very little ability to overcome wrong local minimum value even as new depth frames are obtained.

\subsection{Iterative refinement on real data}

\begin{figure}
\begin{minipage}[t]{0.5\linewidth}
\centering
\includegraphics[height=1.3in]{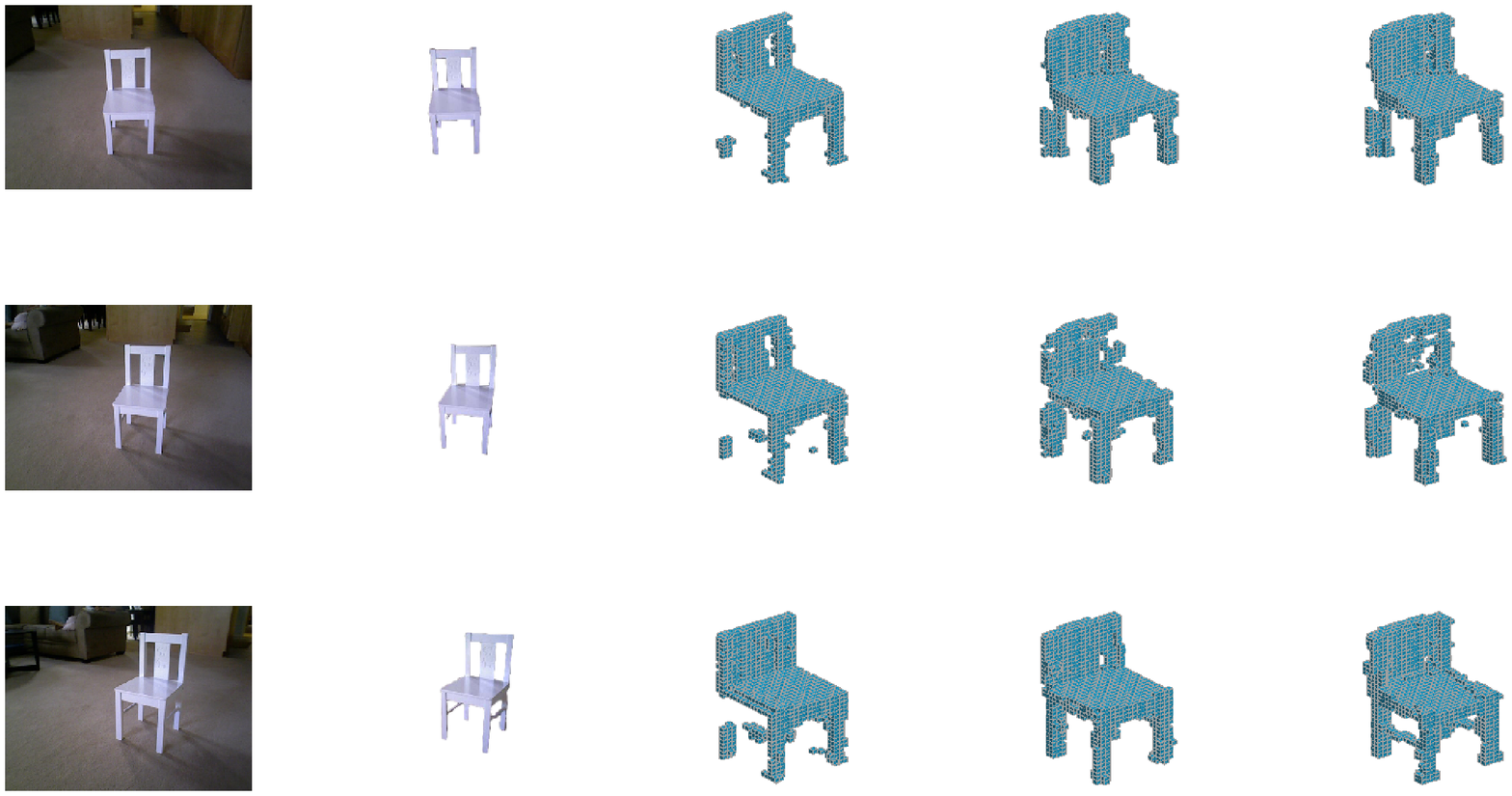}
\end{minipage}%
\begin{minipage}[t]{0.5\linewidth}
\centering
\includegraphics[height =1.3in]{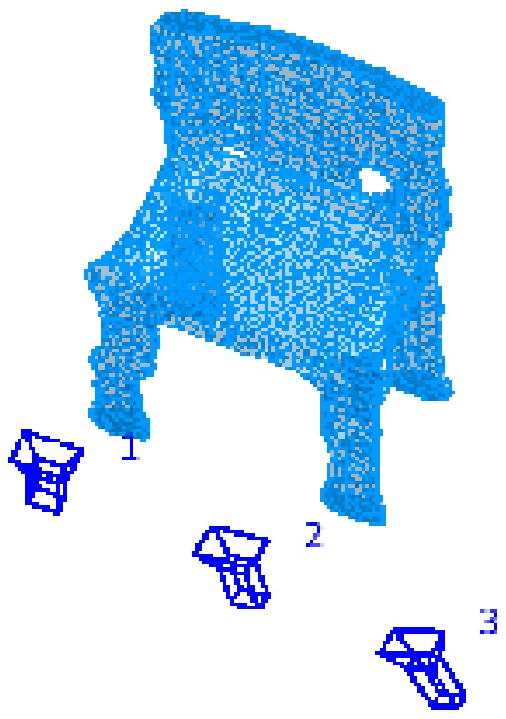}
\end{minipage}
\caption{Result on real data. For the left figure, the first column shows the employed real images of a chair, the second column the segmented chair, the third column the fused measurement occupancy grid, the fourth the result from the auto-encoder, and the last the final result after iterative residual minimization. The right figure also shows a surface estimate of the final result, along with the estimated camera poses.}
  \label{fig-real_cma_au}
\end{figure}

Although the application to real data remains very challenging owing to the segmentation errors and the depth noise, we managed to obtain a few successful results by applying the incremental tracking and mapping pipeline outlined in Section \ref{sec-poserecov} to a few sequences of real Kinect images from the chair dataset \citep{choi2016large}. Figure \ref{fig-real_cma_au} shows one of our obtained results. It confirms that the incremental addition of new frames---each time followed by iterative minimization of the cross-entropy residuals---is able to recover fine structure details compared to the feed-forward result from the auto-encoder alone. Although serving as a basic proof of concept, note that the procedure currently still relies on manual assistance for segmenting the chair (e.g. depth thresholding) and an initial guess for the pose of the first frame.

\section{Discussion}

We regard the present work as a fundamental cornerstone in lifting the parameterization of SLAM in unknown environments to a higher level. We successfully create a marriage between classical but powerful residual minimization techniques and modern deep architectures, able to provide complete and detailed reconstructions at the level of objects even in the light of partially occluded measurements. Despite the employment of residual errors with respect to the measurements, the fact that the estimated shape needs to remain a point in a previously trained latent shape space leads to a good ability to deal with missing data and unmodeled, disturbing influences on the measurements. The present work deals with only a single object, and the world frame coincides with the object frame. We intend to pursue this promising avenue and extend our work to more complex environments with multiple objects of different types. We furthermore intend to introduce a representation that permits joint optimization of poses and geometry.

The community has recently raised interesting questions about the possibility of end-to-end learning in 3D reconstruction. We argue here that from a practical perspective we do not yet have valid incentives to generalize our networks to influences that we already know how to model very well. For example, we started our investigation by using the offspring of \citep{girdhar2016learning}, which generalizes to arbitrary object poses at the input of the network. Our observation is that this significantly increases the dimensionality of the shape space, which in turn affects computational complexity all while reducing the quality of the reconstruction. Our work serves as an example of how to combine valid existing models for poses and residual errors with modern deep architectures, and in particular restrict the generalization domain of the network to parameters for which no explicit model is available (i.e. the intrinsic object shape). By inserting explicit, minimal representations for the well-understood geometric transformations, we achieve a low-dimensional overall parametrization and outstanding performance.

\bibliography{template}

\begin{thebibliography}{31}
\providecommand{\natexlab}[1]{#1}
\providecommand{\url}[1]{\texttt{#1}}
\expandafter\ifx\csname urlstyle\endcsname\relax
  \providecommand{\doi}[1]{doi: #1}\else
  \providecommand{\doi}{doi: \begingroup \urlstyle{rm}\Url}\fi

\bibitem[Wu et~al.(2017)Wu, Wang, Xue, Sun, Freeman, and
  Tenenbaum]{wu2017marrnet}
Jiajun Wu, Yifan Wang, Tianfan Xue, Xingyuan Sun, Bill Freeman, and Josh
  Tenenbaum.
\newblock Marrnet: 3d shape reconstruction via 2.5 d sketches.
\newblock In \emph{Advances In Neural Information Processing Systems}, pages
  540--550, 2017.

\bibitem[Girdhar et~al.(2016)Girdhar, Fouhey, Rodriguez, and
  Gupta]{girdhar2016learning}
Rohit Girdhar, David~F Fouhey, Mikel Rodriguez, and Abhinav Gupta.
\newblock Learning a predictable and generative vector representation for
  objects.
\newblock In \emph{European Conference on Computer Vision}, pages 484--499.
  Springer, 2016.

\bibitem[Yang et~al.(2017)Yang, Wen, Wang, Clark, Markham, and
  Trigoni]{yang20173d}
Bo~Yang, Hongkai Wen, Sen Wang, Ronald Clark, Andrew Markham, and Niki Trigoni.
\newblock 3d object reconstruction from a depth view with adversarial learning.
\newblock \emph{arXiv preprint arXiv:1708.07969}, 2017.

\bibitem[Smith and Meger(2017)]{smith2017improved}
Edward Smith and David Meger.
\newblock Improved adversarial systems for 3d object generation and
  reconstruction.
\newblock \emph{arXiv preprint arXiv:1707.09557}, 2017.

\bibitem[Di et~al.(2016)Di, Dahyot, and Prasad]{di2016deep}
Xinhan Di, Rozenn Dahyot, and Mukta Prasad.
\newblock Deep shape from a low number of silhouettes.
\newblock In \emph{European Conference on Computer Vision}, pages 251--265.
  Springer, 2016.

\bibitem[Yan et~al.(2016)Yan, Yang, Yumer, Guo, and Lee]{yan2016perspective}
Xinchen Yan, Jimei Yang, Ersin Yumer, Yijie Guo, and Honglak Lee.
\newblock Perspective transformer nets: Learning single-view 3d object
  reconstruction without 3d supervision.
\newblock In \emph{Advances in Neural Information Processing Systems}, pages
  1696--1704, 2016.

\bibitem[Rezende et~al.(2016)Rezende, Eslami, Mohamed, Battaglia, Jaderberg,
  and Heess]{rezende2016unsupervised}
Danilo~Jimenez Rezende, SM~Ali Eslami, Shakir Mohamed, Peter Battaglia, Max
  Jaderberg, and Nicolas Heess.
\newblock Unsupervised learning of 3d structure from images.
\newblock In \emph{Advances In Neural Information Processing Systems}, pages
  4996--5004, 2016.

\bibitem[Salas-Moreno et~al.(2014)Salas-Moreno, Glocken, Kelly, and
  Davison]{salas2014dense}
Renato~F Salas-Moreno, Ben Glocken, Paul~HJ Kelly, and Andrew~J Davison.
\newblock Dense planar slam.
\newblock In \emph{Mixed and Augmented Reality (ISMAR), 2014 IEEE International
  Symposium on}, pages 157--164. IEEE, 2014.

\bibitem[Micusik and Wildenauer(2015)]{micusik2015descriptor}
Branislav Micusik and Horst Wildenauer.
\newblock Descriptor free visual indoor localization with line segments.
\newblock In \emph{Computer Vision and Pattern Recognition (CVPR), 2015 IEEE
  Conference on}, pages 3165--3173. IEEE, 2015.

\bibitem[Civera et~al.(2011)Civera, G{\'a}lvez-L{\'o}pez, Riazuelo, Tard{\'o}s,
  and Montiel]{civera2011towards}
Javier Civera, Dorian G{\'a}lvez-L{\'o}pez, Luis Riazuelo, Juan~D Tard{\'o}s,
  and JMM Montiel.
\newblock Towards semantic slam using a monocular camera.
\newblock In \emph{Intelligent Robots and Systems (IROS), 2011 IEEE/RSJ
  International Conference on}, pages 1277--1284. IEEE, 2011.

\bibitem[Salas-Moreno et~al.(2013)Salas-Moreno, Newcombe, Strasdat, Kelly, and
  Davison]{salas2013slam++}
Renato~F Salas-Moreno, Richard~A Newcombe, Hauke Strasdat, Paul~HJ Kelly, and
  Andrew~J Davison.
\newblock Slam++: Simultaneous localisation and mapping at the level of
  objects.
\newblock In \emph{Computer Vision and Pattern Recognition (CVPR), 2013 IEEE
  Conference on}, pages 1352--1359. IEEE, 2013.

\bibitem[Gupta et~al.(2015)Gupta, Arbel{\'a}ez, Girshick, and
  Malik]{gupta2015aligning}
Saurabh Gupta, Pablo Arbel{\'a}ez, Ross Girshick, and Jitendra Malik.
\newblock Aligning 3d models to rgb-d images of cluttered scenes.
\newblock In \emph{Computer Vision and Pattern Recognition (CVPR), 2015 IEEE
  Conference on}, pages 4731--4740. IEEE, 2015.

\bibitem[G{\'a}lvez-L{\'o}pez et~al.(2016)G{\'a}lvez-L{\'o}pez, Salas,
  Tard{\'o}s, and Montiel]{galvez2016real}
Dorian G{\'a}lvez-L{\'o}pez, Marta Salas, Juan~D Tard{\'o}s, and JMM Montiel.
\newblock Real-time monocular object slam.
\newblock \emph{Robotics and Autonomous Systems}, 75:\penalty0 435--449, 2016.

\bibitem[Mu et~al.(2016)Mu, Liu, Paull, Leonard, and How]{mu2016slam}
Beipeng Mu, Shih-Yuan Liu, Liam Paull, John Leonard, and Jonathan~P How.
\newblock Slam with objects using a nonparametric pose graph.
\newblock In \emph{Intelligent Robots and Systems (IROS), 2016 IEEE/RSJ
  International Conference on}, pages 4602--4609. IEEE, 2016.

\bibitem[Koppula et~al.(2011)Koppula, Anand, Joachims, and
  Saxena]{koppula2011semantic}
Hema~S Koppula, Abhishek Anand, Thorsten Joachims, and Ashutosh Saxena.
\newblock Semantic labeling of 3d point clouds for indoor scenes.
\newblock In \emph{Advances in neural information processing systems}, pages
  244--252, 2011.

\bibitem[St{\"u}ckler et~al.(2015)St{\"u}ckler, Waldvogel, Schulz, and
  Behnke]{stuckler2015dense}
J{\"o}rg St{\"u}ckler, Benedikt Waldvogel, Hannes Schulz, and Sven Behnke.
\newblock Dense real-time mapping of object-class semantics from rgb-d video.
\newblock \emph{Journal of Real-Time Image Processing}, 10\penalty0
  (4):\penalty0 599--609, 2015.

\bibitem[Kundu et~al.(2014)Kundu, Li, Dellaert, Li, and Rehg]{kundu2014joint}
Abhijit Kundu, Yin Li, Frank Dellaert, Fuxin Li, and James~M Rehg.
\newblock Joint semantic segmentation and 3d reconstruction from monocular
  video.
\newblock In \emph{European Conference on Computer Vision}, pages 703--718.
  Springer, 2014.

\bibitem[McCormac et~al.(2017)McCormac, Handa, Davison, and
  Leutenegger]{mccormac2017semanticfusion}
John McCormac, Ankur Handa, Andrew Davison, and Stefan Leutenegger.
\newblock Semanticfusion: Dense 3d semantic mapping with convolutional neural
  networks.
\newblock In \emph{Robotics and Automation (ICRA), 2017 IEEE International
  Conference on}, pages 4628--4635. IEEE, 2017.

\bibitem[H\"ane et~al.(2017)H\"ane, Zach, Cohen, and Pollefeys]{haene17}
C~H\"ane, C~Zach, A~Cohen, and M~Pollefeys.
\newblock Dense semantic 3d reconstruction.
\newblock \emph{Transactions on Pattern Analysis and Machine Intelligence
  (TPAMI)}, 2017.

\bibitem[Bloesch et~al.(2018)Bloesch, Czarnowski, Clark, Leutenegger, and
  Davison]{bloesch2018codeslam}
Michael Bloesch, Jan Czarnowski, Ronald Clark, Stefan Leutenegger, and Andrew~J
  Davison.
\newblock Codeslam-learning a compact, optimisable representation for dense
  visual slam.
\newblock \emph{arXiv preprint arXiv:1804.00874}, 2018.

\bibitem[Zhu et~al.(2017)Zhu, Wang, Lin, Wang, and Lucey]{zhu2017semantic}
Rui Zhu, Chaoyang Wang, Chen-Hsuan Lin, Ziyan Wang, and Simon Lucey.
\newblock Semantic photometric bundle adjustment on natural sequences.
\newblock \emph{arXiv preprint arXiv:1712.00110}, 2017.

\bibitem[Fan et~al.(2016)Fan, Su, and Guibas]{FanSG16}
Haoqiang Fan, Hao Su, and Leonidas~J. Guibas.
\newblock A point set generation network for 3d object reconstruction from a
  single image.
\newblock \emph{CoRR}, abs/1612.00603, 2016.

\bibitem[Dame et~al.(2013)Dame, Prisacariu, Ren, and Reid]{dame13}
A~Dame, V~A Prisacariu, C~Y Ren, and I~Reid.
\newblock Dense reconstruction using 3d object shape priors.
\newblock In \emph{IEEE Conference on Computer Vision and Pattern Recognition},
  2013.

\bibitem[Dauphin et~al.(2014)Dauphin, Pascanu, Gulcehre, Cho, Ganguli, and
  Bengio]{dauphin2014identifying}
Yann~N Dauphin, Razvan Pascanu, Caglar Gulcehre, Kyunghyun Cho, Surya Ganguli,
  and Yoshua Bengio.
\newblock Identifying and attacking the saddle point problem in
  high-dimensional non-convex optimization.
\newblock In \emph{Advances in neural information processing systems}, pages
  2933--2941, 2014.

\bibitem[Dai et~al.(2017)Dai, Qi, and Nie{\ss}ner]{dai2017shape}
Angela Dai, Charles~Ruizhongtai Qi, and Matthias Nie{\ss}ner.
\newblock Shape completion using 3d-encoder-predictor cnns and shape synthesis.
\newblock In \emph{Proc. IEEE Conf. on Computer Vision and Pattern Recognition
  (CVPR)}, volume~3, 2017.

\bibitem[Wu et~al.(2016)Wu, Zhang, Xue, Freeman, and Tenenbaum]{wu2016learning}
Jiajun Wu, Chengkai Zhang, Tianfan Xue, Bill Freeman, and Josh Tenenbaum.
\newblock Learning a probabilistic latent space of object shapes via 3d
  generative-adversarial modeling.
\newblock In \emph{Advances in Neural Information Processing Systems}, pages
  82--90, 2016.

\bibitem[Lin et~al.(2013)Lin, Chen, and Yan]{lin2013network}
Min Lin, Qiang Chen, and Shuicheng Yan.
\newblock Network in network.
\newblock \emph{arXiv preprint arXiv:1312.4400}, 2013.

\bibitem[Chang et~al.(2015)Chang, Funkhouser, Guibas, Hanrahan, Huang, Li,
  Savarese, Savva, Song, Su, et~al.]{chang2015shapenet}
Angel~X Chang, Thomas Funkhouser, Leonidas Guibas, Pat Hanrahan, Qixing Huang,
  Zimo Li, Silvio Savarese, Manolis Savva, Shuran Song, Hao Su, et~al.
\newblock Shapenet: An information-rich 3d model repository.
\newblock \emph{arXiv preprint arXiv:1512.03012}, 2015.

\bibitem[Hansen(2016)]{hansen2016cma}
Nikolaus Hansen.
\newblock The cma evolution strategy: A tutorial.
\newblock \emph{arXiv preprint arXiv:1604.00772}, 2016.

\bibitem[Besl and McKay(1992)]{besl1992method}
Paul~J Besl and Neil~D McKay.
\newblock Method for registration of 3-d shapes.
\newblock In \emph{Sensor Fusion IV: Control Paradigms and Data Structures},
  volume 1611, pages 586--607. International Society for Optics and Photonics,
  1992.

\bibitem[Choi et~al.(2016)Choi, Zhou, Miller, and Koltun]{choi2016large}
Sungjoon Choi, Qian-Yi Zhou, Stephen Miller, and Vladlen Koltun.
\newblock A large dataset of object scans.
\newblock \emph{arXiv preprint arXiv:1602.02481}, 2016.

\end{thebibliography}
\end{document}